\documentclass{article}
\usepackage{spconf,amsmath,graphicx,hyperref}

\usepackage{algorithm}
\usepackage{algorithmic}

\usepackage{amsmath}
\usepackage{amssymb}
\usepackage{amsfonts}

\title{LyriCAR: A Difficulty-Aware Curriculum Reinforcement Learning Framework for Controllable Lyric Translation}
\name{Le Ren$^{1}$, Xiangjian Zeng$^{2}$, Qingqiang Wu$^{1,3,4,5,6}$, Ruoxuan Liang$^{3}$}
\address{
    $^{1}$School of Informatics, Xiamen University \\
    $^{2}$School of Journalism and Communication, Xiamen University  \\
    $^{3}$School of Film, Xiamen University \\
    $^{4}$Xiamen Key Laboratory of Intelligent Storage and Computing, Xiamen University \\ $^{5}$Key Laboratory of Digital Protection and Intelligent Processing of Intangible Cultural Heritage of \\ Fujian and Taiwan, Ministry of Culture and Tourism, Xiamen University \\ $^{6}$Institute of Artificial Intelligence, Xiamen University\
}

\begin{document}
%\ninept
%
\maketitle
\begin{abstract}

Lyric translation is a challenging task that requires balancing multiple musical constraints. Existing methods often rely on hand-crafted rules and sentence-level modeling, which restrict their ability to internalize musical–linguistic patterns and to generalize effectively at the paragraph level, where cross-line coherence and global rhyme are crucial. In this work, we propose \textbf{LyriCAR}, a novel framework for  controllable lyric translation that operates in a fully unsupervised manner. LyriCAR introduces a difficulty-aware curriculum designer and an adaptive curriculum strategy, ensuring efficient allocation of training resources, accelerating convergence, and improving overall translation quality by guiding the model with increasingly complex challenges. Extensive experiments on the EN–ZH lyric translation task show that LyriCAR achieves state-of-the-art results across both standard translation metrics and multi-dimensional reward scores, surpassing strong baselines. Notably, the adaptive curriculum strategy reduces training steps by nearly 40\% while maintaining superior performance. Code, data and model can be accessed at \texttt{https://github.com/rle27/LyriCAR}.
\end{abstract}
\section{Introduction}
\label{sec:intro}

Lyric translation is worth studying because language barriers can limit the enjoyment of global music. However, unlike neural machine translation tasks, lyric translation emphasizes musicality preservation, which means handling rhyme, rhythm, and semantic quality simultaneously, requiring a balance across often conflicting dimensions.

To achieve multi-dimensional, controllable lyrics translation, extensive research has been conducted. \cite{GagaST} enforces acoustic-linguistic alignment during decoding, penalizing candidate sequences violate alignment rules during the beam search process; \cite{LTAG} directly injects melody and alignment ratio into the input of the Transformer encoder and designs a lightweight alignment decoder to to predict monotonic lyric-melody alignment; \cite{control} encodes rhythm-related constraints (length, rhyme, word boundaries) as control tokens; and \cite{narrate} trains a reward model to jointly optimize singability and translation quality through reinforcement learning.

Despite their innovations, existing approaches exhibit several fundamental limitations: (1) an overreliance on manually engineered constraints or heuristic decoding strategies, rather than endowing the model with the capacity to internalize music–language regularities \cite{GagaST,control}; (2) narrow coverage of constraint dimensions \cite{LTAG} and labor-intensive constraint annotations\cite{control}; (3) inadequate paragraph-level modeling, with sentence-level frameworks failing to capture cross-line rhyme patterns \cite{control}, and search-based methods incurring prohibitive computational complexity that undermines real-time applicability \cite{narrate}; and (4) suboptimal data utilization, relying on coarse curriculum learning\cite{LTAG} or weakly aligned text–melody pairs, leaving the majority of training data semantically disconnected from musical structure.

\begin{figure*}[th]
    \centering
    \includegraphics[width=1\linewidth]{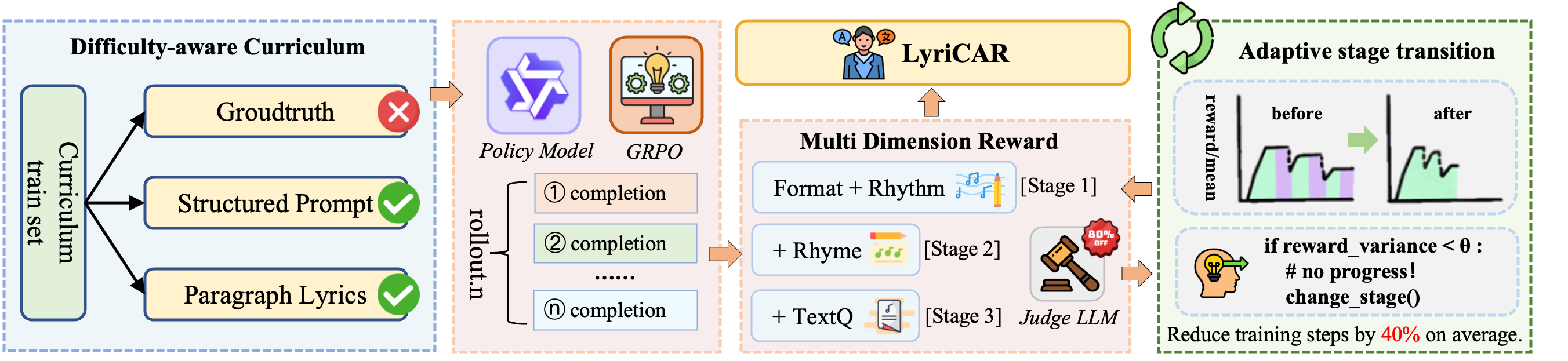}
    \caption{Pipeline of LyriCAR}
    \label{fig:pipeline}
\end{figure*}

Building upon the limitations of prior work, we aim to address the challenge of lyric translation in a holistic manner internalizing the interplay between linguistic fidelity and musicality through curriculum based reinforcement learning.

Our key contributions are:
\begin{itemize}
    \item We design a difficulty-aware curriculum strategy combined with staged structural cues, enabling multi-dimensional lyric translation without target lyric or alignment annotations.
    \item We propose \textbf{LyriCAR}, a adaptive reinforcement learning framework that jointly optimizes semantic and musical dimensions, achieving efficient, end-to-end translation.
    \item Extensive experiments demonstrate state-of-the-art performance, surpassing baselines in automatic metrics and multi-dimensional reward scores, while reducing training steps by nearly 40\%.
\end{itemize}

\section{Methodology}
\label{sec:format}

As illustrated in Fig. 1, our framework consists of three modules: (1) a difficulty-aware curriculum designer; (2) a reinforcement learning method with multi-dimensional reward guided reward; and (3) a convergence guided adaptive curriculum strategy.

% Dynamic Adaptive Training Acceleration Method
% based on real-time reward convergence

\subsection{Difficulty-Aware Curriculum Designer}
\label{ssec:subhead}

To achieve fast convergence and high learning efficiency within a fully alignment-free and annotation-free unsupervised setting, we propose a difficulty-aware curriculum designer that relies solely on raw paragraph-level source lyrics. The intrinsic linguistic complexity of each paragraph is quantified BERT-based perplexity \cite{salazar-etal-2020-masked} scoring and lexicon-based linguistic complexity features inspired by LIWC \cite{pennebakerlinguistic}, capturing lexical diversity, syntactic depth, and rhyme density. Based on these measures, the dataset is stratified into three levels of difficulty, namely Easy, Medium, and Hard, which are sampled in a staged and progressively challenging manner, as shown in Table \ref{tab:sample_tier}, to construct the training sets for three successive stages. In this way, we establish the first truly unsupervised, end-to-end framework for paragraph-level lyric translation, effectively overcoming both the data scarcity and the structural limitations that have hindered previous approaches.

\subsection{Reinforcement Learning with Multi-Dimensional Reward}
\label{ssec:2.2}

As showed in Fig \ref{fig:pipeline}, we fine-tune the large language model Qwen3-8B\cite{qwen3}, which provides a solid foundation for the subsequent lyric translation task. To distinguish high-quality from suboptimal translations, candidate completions are evaluated using four reward functions that capture constraints across different dimensions:

Format compliance ($R_{fmt}$): Ensures that special tokens marking sentence boundaries within a paragraph are preserved.

\begin{equation}
\mathrm{Format}(S) = 1 - \frac{\sum_{i=1}^{N} |L_i - \hat{L}|}{N \cdot \hat{L}} \label{eq:format}
\end{equation}

Rhythm compliance ($R_{rtm}$): Ensures output length matches target syllable count.

\begin{equation} \label{eq:rhythm}
\mathrm{Rhyme}(S) = \frac{1}{N-1} \sum_{i=1}^{N-1} \mathrm{sim}(\sigma_i, \sigma_{i+1}) 
\end{equation} 

Rhyme compliance ($R_{rym}$): Encourages consistent rhyme patterns across sentences within a paragraph.

\begin{equation}
\mathrm{Rhythm}(S) = 1 - \frac{\sum_{i=1}^{M} |d_i - \hat{d}_i|}{\sum_{i=1}^{M} \hat{d}_i}
\end{equation}

Text quality compliance ($R_{txtQ}$): Ensures that the translation faithfully conveys the original cultural and semantic content.

\begin{equation}
\mathrm{TextQuality}(S) = \{-1,0,1\}
\end{equation}

Motivated by \cite{judge,judge2}, text quality is evaluated via a prompt-based Judge LLM, which maps categorical judgments to discrete scores. Only ambiguous samples with scores between 0.5 and 0.7 are scored, reducing computation by roughly 80\%. These reward signals are combined through weighted summation to form the final reward score:

\begin{equation}
    \begin{split}
        \mathrm{Reward}(S) &= \lambda_1 \cdot R_{fmt} 
        + \lambda_2 \cdot R_{rtm} \\
        & + \lambda_3 \cdot R_{rym} 
        + \lambda_4 \cdot R_{txtQ}
    \end{split}
\end{equation}

where $\lambda_i$ represents the weight of each component.

Rather than being applied as external penalties, these reward signals are learned internally by the model via Group Relative Policy Optimization (GRPO)\cite{GRPO}. GRPO operates within a reinforcement learning framework, comparing candidate outputs in groups to compute relative advantages. Specifically, for a group $g$ of candidate completions with total reward $R$, the group-relative advantage for candidate $k$ is defined as:

\begin{equation}
A^{(g)}_k = R_k - \frac{1}{|g|} \sum_{j \in g} R_j
\end{equation}

The policy $\pi_\theta$ is then updated to maximize the expected group-relative advantage across sampled groups:

\begin{equation} \label{eq:grpo}
\mathcal{L}(\theta) = - \mathbb{E}_{g \sim \mathcal{G}, k \sim \pi_\theta} \left[ \log \pi_\theta(k) \, A^{(g)}_k \right]
\end{equation}

This approach allows the model to autonomously learn the trade-offs between conflicting objectives, such as semantic fidelity, rhythm, rhyme, and text quality, without relying on handcrafted rules or external penalties.

\begin{algorithm}
    \caption{Reward Convergence Guided Curriculum Adaptation Strategy}
    \label{alg:dynamic_curriculum}
    \begin{algorithmic}[1]
        \REQUIRE Initial policy $\pi_0$, curriculum stages $\{C_1, ..., C_N\}$,  
                 reward variance threshold $\tau$, patience $k$, interval $I$
        \ENSURE Final policy $\pi^*$
        
        \STATE Initialize $\pi \gets \pi_0$, stage $i \gets 1$, dataset $D \gets C_i$
        
        \WHILE{$i \leq N$}
            \STATE Train $\pi$ on $D$ using GRPO (Eq.~\ref{eq:grpo})
            \IF{every $I$ epochs}
                \STATE Record validation reward $\bar{R}_t$ in sliding window $W$
                \IF{$|W| \geq k$ and $\mathrm{Var}(W) < \tau$}
                    \STATE $i \gets i+1$, $D \gets C_i$, reset $W$
                \ENDIF
            \ENDIF
        \ENDWHILE
        
        \RETURN $\pi$
    \end{algorithmic}
\end{algorithm}

\subsection{Convergence Guided Adaptive Curriculum Strategy}

In practical training, we observe heterogeneous convergence across curriculum stages: early tasks with simple prompts are mastered rapidly, while later stages involving multi-dimensional constraints tend to stagnate. To mitigate this imbalance, we adopt a reward-convergence-guided stage adaptation mechanism, as shown in Algorithm~\ref{alg:dynamic_curriculum}. Our design is grounded in curriculum learning principles\cite{curri,curri2} and self-paced learning\cite{selfpace}, which advocate presenting progressively harder data to accelerate convergence and improve final performance.

Building on competence based curriculum \cite{competence} and teacher–student strategies\cite{teacher}, we monitor reward trajectories and employ a sliding-window variance criterion to detect saturation. Once the variance remains below a threshold $\theta$ (initialized from a small validation study and subsequently fixed), the model transitions to the next stage, thereby introducing more challenging data and richer reward dimensions.

This adaptive scheduling not only prevents overfitting in early stages and under-exploration in later ones, but also aligns training progress with the model’s actual learning dynamics. By reallocating resources to harder tasks exactly when simpler ones are sufficiently mastered, the mechanism reduces wasted computation, minimizes reliance on manual hyperparameter tuning, and improves both efficiency and stability across diverse experimental settings.

\section{Experiments}
\label{sec:pagestyle}

\subsection{Experimental configuration}

\subsubsection{Datasets and metrics}

The dataset is derived from DALI\cite{DALI}, a large collection of synchronized audio, lyrics, and vocal notes. From 6,984 English songs after filtering, we extracted lyrics and constructed 9,600 paragraphs per stage, as summarized in Table~\ref{tab:sample_tier}.

\begin{table}[hb]
    \centering
    \begin{tabular}{c|c c c|c}
    \hline
         & Easy & Medium & Hard & Paragraphs \\
    \hline
    Stage1 & 0.5 & 0.3 & 0.2 & 9600 \\
    Stage2 & 0.3 & 0.5 & 0.2 & 9600 \\
    Stage3 & 0.2 & 0.3 & 0.5 & 9600 \\
    \hline
    \end{tabular}
    \caption{Difficulty distribution and dataset size of each stage}
    \label{tab:sample_tier}
\end{table}

\begin{table*}[thb]
    \centering
    \begin{tabular}{c|c c|c c c c|c}
        \hline
         & BLUE & COMET & RhymeFreq & RhythmS & TransQ & Sum & TrainTime \\
        \hline
        Ou et al. (2023)\cite{control} & 18.01 & 71.94 & - & - & - & - & - \\
        Ye et al. (2024)\cite{narrate} & 18.80 & 74.14 & - & - & - & - & - \\
        Qwen3-8B-base & 16.87 & 77.37 & 0.42 & 0.48 & 0.55 & 1.45 & - \\
        LyriCAR-F & 20.45 & 79.82 & 0.58 & 0.63 & 0.71 & 0.5 & 100\% \\
        LyriCAR-SS & 21.02 & 80.37 & 0.61 & 0.66 & 0.74 & 0.6 & 85\% \\
        LyriCAR-SD & \textbf{21.37} & \textbf{81.12} & \textbf{0.65} & \textbf{0.70} & \textbf{0.77} & \textbf{0.7} & \textbf{51\%} \\
        \hline
    \end{tabular}
    \caption{Comparisons with previous SOTA, where LyriCAR-F means the full-data versiom, LyriCAR-SS and LyriCAR means staged staic version and staged dynamic version seperately.}
    \label{tab:mainresult}
\end{table*}

\subsubsection{Experimental setup}
\label{sssec:subhead}

All experiments were conducted on 8 NVIDIA A800 80GB GPUs. Training was initialized from the pretrained Qwen3-8B model, and hyperparameters were adjusted progressively across curriculum stages. Specifically, learning rates were set to $1\times10^{-6}$, $5\times10^{-7}$, and $1\times10^{-7}$ for increasing difficulty levels, while the KL loss coefficient was correspondingly scheduled as 0.01, 0.05, and 0.1. The batch size was fixed at 128, with a PPO mini-batch size of 64 and a PPO micro-batch size per GPU of 16.

\subsection{Main Results}

To ensure comprehensive evaluation, we assess performance on both supervised and unsupervised settings. On the parallel test set, BLEU\cite{bleu} and COMET\cite{comet} capture translation quality under supervised metrics, while on the unlabeled DALI validation subset, the multi-dimensional reward score (§\ref{ssec:2.2}) provides unsupervised evaluation. This complementary setup offers a balanced view, with results summarized in Table~\ref{tab:mainresult}.

LyriCAR achieves state-of-the-art performance across all automatic metrics (BLEU, COMET) on the EN–CH lyric translation task, outperforming all baselines—including the strong Qwen3-8B model. It also obtains the highest scores on our multi-dimensional reward evaluation, indicating that the model has truly internalized musical-linguistic alignment patterns rather than relying on shallow correlations. Notably, LyriCAR-SD further reduces training steps by 34\% compared to LyriCAR-SS while delivering superior translation quality.

These results demonstrate that our method not only meets the demanding requirements of lyric translation but also effectively balances high-quality generation with computational efficiency. Importantly, this is achieved without reliance on large-scale parallel data or costly manual annotations, highlighting the practicality and scalability of the proposed framework.

\begin{figure}[htb]
\begin{minipage}[a]{.48\linewidth}
  \centering
  \centerline{\includegraphics[width=4.0cm]{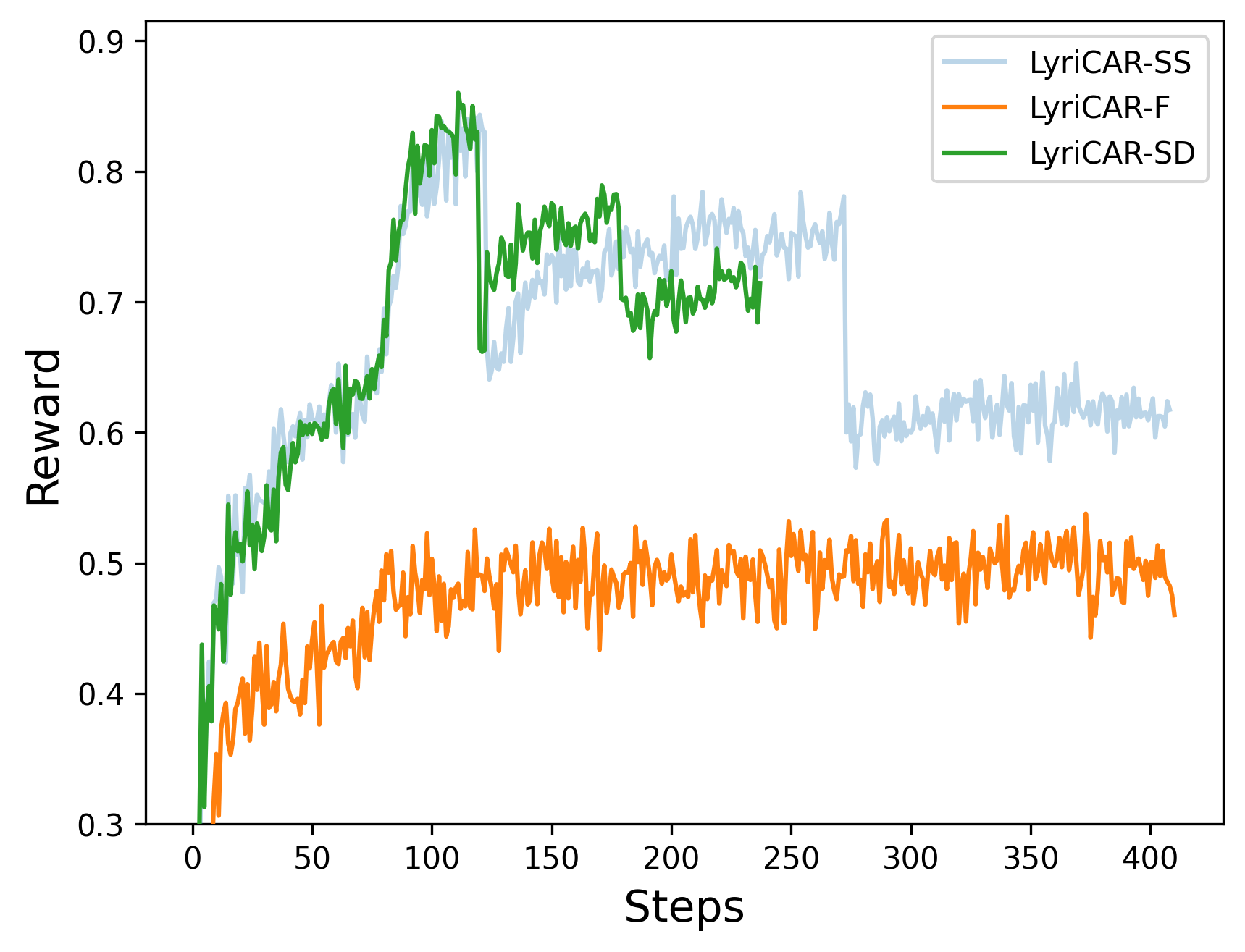}}
%  \vspace{1.5cm}
  \centerline{(a) Reward trajectories}\medskip
\end{minipage}
\hfill
\begin{minipage}[a]{0.48\linewidth}
  \centering
  \centerline{\includegraphics[width=4.0cm]{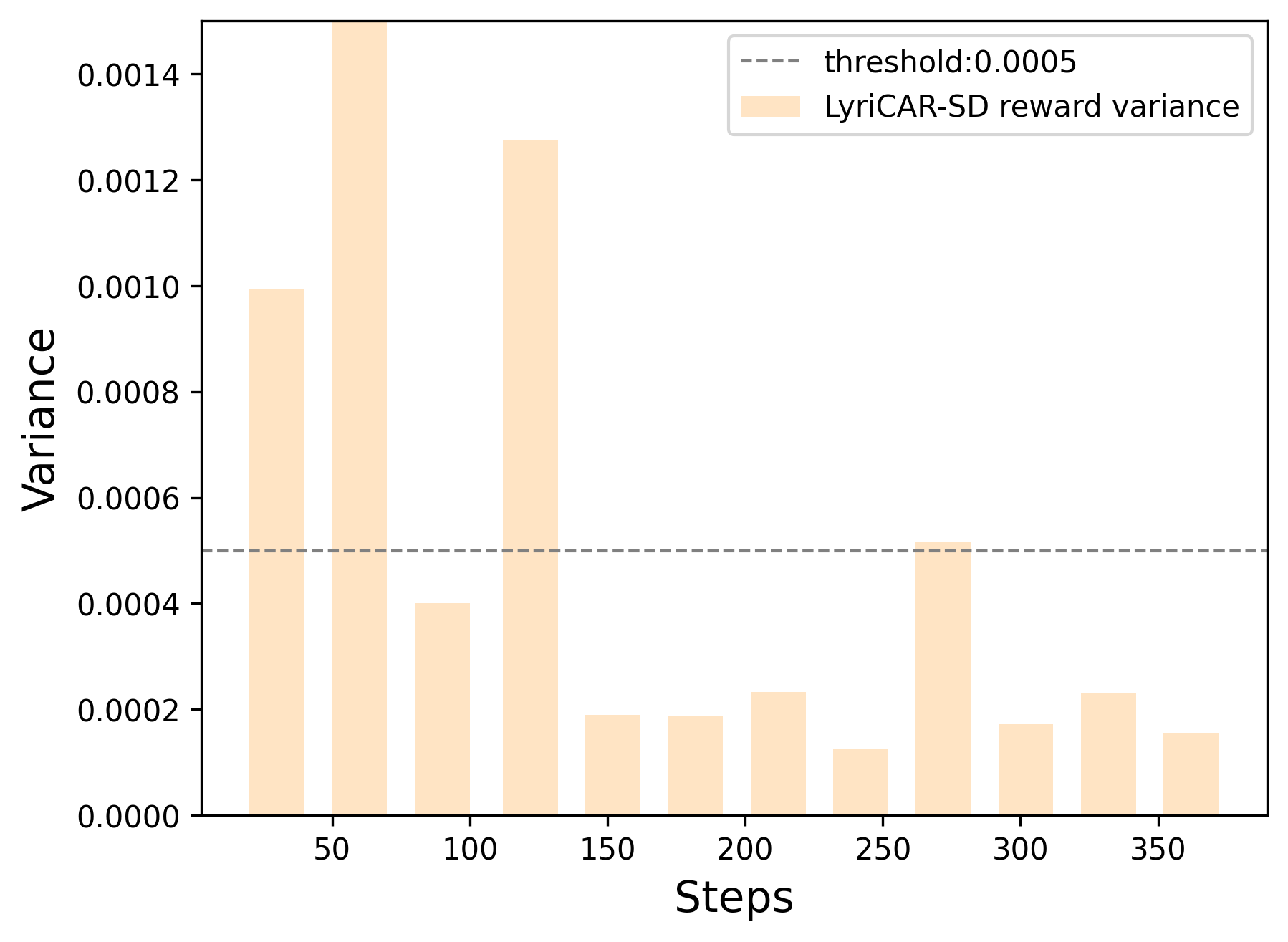}}
%  \vspace{1.5cm}
  \centerline{(b) Reward variance(LyriCAR-SD)}\medskip
\end{minipage}
\caption{Ablation study results.}
\label{fig:res}
\end{figure}

\subsection{Ablation study}

We conducted training under two settings: full-data training and curriculum-based training, ensuring identical total data volume and number of epochs across both paradigms. As shown in Fig.~\ref{fig:res}(a), the performance of full-data training eventually oscillates around 0.5, which coincides with the lower bound imposed by the Judge LLM. This indicates that full-data training places a substantial burden on the model, hindering its ability to efficiently acquire complex musical-linguistic patterns and limiting further performance gains.

In contrast, curriculum-based training (LyriCAR-SS) enables rapid improvement in foundational capabilities during early stages. Although performance naturally dips at stage transitions due to the increased difficulty of tasks, the final results stabilize around 0.6 which is significantly higher than full-data training.

Moreover, when combined with the Reward-Convergence-Guided Curriculum Adaptation strategy, LyriCAR-SD not only boosts the final performance to approximately 0.7 but also reduces the required training steps by 40\%.

These ablation results demonstrate the effectiveness of our curriculum design and adaptive stage-switching mechanism. By aligning training effort with learning dynamics, our approach achieves superior performance with improved learning efficiency, confirming its advantage over static training paradigms. Compared to conventional full-data training, our method strikes a better balance between efficiency and robustness. This validates the central premise of LyriCAR: that dynamically guided curricula can effectively internalize multi-dimensional musical–linguistic patterns, leading to both higher translation quality and more economical use of computational resources.

\section{Conclusion}

We propose LyriCAR, a fully unsupervised framework for multi-dimensional lyric translation that simultaneously balances rhythm, rhyme, and text quality. Our approach combines a difficulty-aware curriculum with a reward convergence guided stage adaptive strategy. Unlike prior approaches that rely heavily on engineered constraints or sentence-level modeling, LyriCAR enables the model to internalize the underlying principles of translation and extend them to paragraph-level generation. The framework achieves state-of-the-art results across multiple evaluation dimensions, while reducing training steps by approximately 40\% compared with strong baselines. These findings highlight LyriCAR as a robust and generalizable solution for cross-lingual music translation, laying the groundwork for future research in musically informed language generation.

\section{Acknowlegement}

This work is supported by the Solfeggio ear training intelligent robot and cloud platform research and development project for music education (No.2024CXY0102), the 3D visualization digital twin integrated control system (No.2023C\\XY0111), the public technology service platform project of Xiamen City (No.3502Z20231043) and Fujian Provincial Science and Technology Major Project (No.
2024HZ022003).

%\vfill
%\pagebreak

\vfill\pagebreak

\bibliographystyle{IEEEbib}
\bibliography{strings,refs}

\end{document}